\documentclass[12pt]{iopart}
\usepackage{color}
\usepackage{iopams} 
\usepackage{subfigure}
\usepackage{epsfig}
\usepackage{graphicx}
\usepackage{epstopdf}
\expandafter\let\csname equation*\endcsname\relax
\expandafter\let\csname endequation*\endcsname\relax
\usepackage{amsmath,amssymb}    
\usepackage[numbers]{natbib}
\usepackage[titletoc,toc,title]{appendix}
\usepackage{graphicx}
\usepackage{subfig}
\usepackage{caption}
\usepackage{sidecap}
\usepackage[applemac]{inputenc}
\usepackage{natbib}
\usepackage{graphicx}           
\usepackage{flafter}            
\usepackage{amsmath,amssymb}    
\usepackage{bm}                 
\usepackage{memhfixc}           
\usepackage[activate]{pdfcprot} 
\usepackage{pdfsync}            

\usepackage{verbatim}

\newcommand{\new}{\color{black}}
\usepackage{algorithm,algcompatible,amsmath}
\algnewcommand\INPUT{\item[\textbf{Input:}]}%
\algnewcommand\OUTPUT{\item[\textbf{Output:}]}%
\begin{document}
\title{Analysis of Bayesian Inference Algorithms by the Dynamical Functional Approach}
\vspace{0.1pc}
\author{ Burak \c{C}akmak and Manfred Opper }
\vspace{0.3pc}
\address {Department of Artificial Intelligence, Technische Universit\"{a}t Berlin, Marchstraße 23, $~~$Berlin 10587, Germany}
\eads{\mailto{\{burak.cakmak, manfred.opper\}@tu-berlin.de}}

\begin{abstract}
	We analyze the dynamics of an algorithm for approximate inference with large Gaussian latent variable models in a student-teacher scenario. To model nontrivial dependencies between the latent variables, we assume random covariance matrices drawn from rotation invariant ensembles. For the case of perfect data-model matching, the knowledge of static order parameters derived from the replica method allows us to obtain efficient algorithmic updates in terms of matrix-vector multiplications with a fixed matrix. Using the dynamical functional approach, we obtain an exact effective stochastic process in the thermodynamic limit for a single node. From this, we obtain closed-form expressions for the rate of the convergence. Analytical results are excellent agreement with simulations of single instances of large models.
\end{abstract}

\vspace{1pc}




\bibliographystyle{plain}
\def\mathlette#1#2{{\mathchoice{\mbox{#1$\displaystyle #2$}}%
                               {\mbox{#1$\textstyle #2$}}%
                               {\mbox{#1$\scriptstyle #2$}}%
                               {\mbox{#1$\scriptscriptstyle #2$}}}}
\newcommand{\matr}[1]{\mathlette{\boldmath}{#1}}
\newcommand{\RR}{\mathbb{R}}
\newcommand{\CC}{\mathbb{C}}
\newcommand{\NN}{\mathbb{N}}
\newcommand{\ZZ}{\mathbb{Z}}
\newcommand{\at}[2][]{#1|_{#2}}
\def\oneh{\frac{1}{2}}
\newcommand{\Kb}{{\mathbf{K}}}
\vspace{2pc}

\section{Introduction}
Tools of statistical mechanics have been extensively used in the late 1980's and the early
1990's to analyze the learning properties of large neural networks and related learning models \cite{gardner1988space,watkin1993statistical,opper1996statistical,nishimori2001statistical}.
The powerful combination of statistical ensembles of learning machines with the application of the replica approach has allowed  for the exact computation of average case learning 
performance. Remarkably, this could be achieved without having to deal with the details 
of concrete learning algorithms. Unfortunately, since much of the research was 
somewhat disconnected from practical existing machine learning approaches, this seemingly advantage has also led to a decline of the research activities in the field in the later 1990s. 

More recently, the situation has again changed considerably. The establishment of relations between advanced mean field approximations (related to the so-called cavity method) and message passing algorithms for graphical statistical  models has led to novel interest in the interdisciplinary area of statistical mechanics of learning and inference. Message passing algorithms were not only found (under certain  conditions) to achieve optimal performance for large systems, but could also be analyzed dynamically by {\em density evolution} techniques \cite{richardson2001design,Bolthausen}. While most of this research concentrated originally on sparse networks \cite{richardson2001design}, there was a growing interest on studying inference with networks of densely coupled probabilistic units \cite{tanaka2005approximate,Bayati,krzakala2012statistical,bayati2015universality}. By taking formally the large density limit in the theory of message passing  one obtains so-called AMP (approximate 
message passing) algorithms \cite{Kabashima,Donoha,Rangan} which have been applied to a great variety of models. Fixed points of the AMP algorithm were found to coincide with the solutions of 
corresponding TAP (Thouless-Anderson-Palmer) mean field equations for  
statistical averages of the stochastic nodes. This heuristic approach might be criticized because certain independence assumptions made in the cavity approach may not be valid for dense networks. Nevertheless, exactness of the final results could be established for certain simple distributions of random network couplings. To go beyond such simple distributions, more complex ensembles allowing for dependencies 
between couplings in dense systems could be treated within an adaptive TAP approach \cite{Adatap} motivated by earlier work on spin glasses \cite{Mezard}. This research had (in parts) led to the development of the {\em expectation propagation} (EP) algorithms \cite{Minka1} in the field of machine learning. Assuming that matrices of 
network couplings are realizations of rotation invariant ensembles, prediction 
properties of the EP-style VAMP (vector-AMP) algorithms could be analyzed rigorously by a density evolution method using methods of random matrix theory \cite{VAMP,takeuchi2017rigorous}.
The density evolution approach so far deals mainly with the computation of the temporal development of predictions errors which are derived from equal time marginals of the 
distribution of trajectories of an algorithm's dynamics. It would be important to extend these results to multivariate statistical properties of trajectories which allow for more detailed computations of the algorithm's performance including the convergence speed towards the fixed point. 

In this paper, we will present such an approach. Based on our previous studies on 
the dynamics of solving TAP equations for Ising spin systems \cite{Opper16,CakmakOpper19},  we introduce a VAMP-style algorithm for inference in {\em Gaussian latent} probabilistic models. These are 
important statistical data models with applications to classification and regression.
We use the method of {\em dynamical functional theory} (DFT) to study the average case properties of algorithms in the large system limit. DFT is  a statistical mechanics tool for analyzing dynamical systems \cite{Martin} based on partition functions over trajectories. From these, effective distributions of entire trajectories for a {\em single node} can be obtained. In the past, the method was mainly
applied to the dynamics of spin-glasses \cite{Eisfeller} and neural networks \cite{sompolinsky1988chaos} with random independent couplings. We also refer to the study \cite{Mimura} where the DFT was used to analyze AMP-style algorithms (in the context of communication theory) with again random independent 
couplings assumption.

The novel contributions presented in this paper are twofold: 
On a technical level we extend the DFT approach of \cite{CakmakOpper19}
to a combination of a teacher-student scenario for the generation of data 
together with the assumption of arbitrary rotation invariant random matrix couplings.
From a more practical machine learning point of view, our new algorithm 
is designed for the case, where the class of data generating models is assumed to be known up to its parameters. This means we neglect model mismatch. In this case, we can use the knowledge of 
equilibrium order parameters given by the replica method to a obtain a simplified algorithm with a significant reduction of computational complexity.

The paper is organized as follows: In Section 2 we introduce the probabilistic model considered for the Bayesian inference. Section 3 provides a brief presentation on the TAP equations. In Section 4 we present our new algorithm for solving the TAP equations and in Section 5 we study its thermodynamic  properties using the method of DFT. Comparisons of the theory with simulations are given in
Section 6. Section 7 presents a summary and outlook.
The derivations of our results are located at the Appendix.
 
\section{Models with Gaussian latent variables}
We consider the problem of approximate Bayesian inference for posterior distributions over the Gaussian latent vector $\matr \theta\in\Re^{N\times 1}$ of the type 
\begin{equation}
p(\matr \theta \vert\matr y,\matr K)\doteq\frac 1 Z e^{-\frac 1 2\matr \theta ^\top\matr K^{-1}\matr \theta}\prod_{i\leq N}p(y_i \vert \theta_i)\label{Gausslate}
\end{equation}
where $Z$ is a normalization constant. This model assumes that the components 
of the vector  $\matr y$ 
of $N$ real data values  are generated independently from a likelihood  $p(y \vert \theta)$
based on a vector of unknown parameters $\matr \theta$. Prior statistical knowledge about 
$\matr \theta$ is introduced by the correlated Gaussian with covariance $\matr K$\footnote{When $\matr K $ is singular $\matr K ^{-1}$ stands for the pseudo inverse of $\matr K$.} A well known example for (\ref{Gausslate}) is the problem of Bayesian learning 
of a noisy perceptron---also known as {\em probit regression} \cite{neal1997monte}---for which we assume binary class labels $y_i = \pm 1$. For more details on the standard interpretation of the noisy perceptron, see section \ref{Sim}.

Our concern is to design and analyze an iterative algorithm that (approximately) computes the vector of posterior means
\begin{equation}
\matr m\doteq \mathbb E[\matr \theta], \quad \matr \theta \sim p(\matr \theta \vert\matr y,\matr K)
\end{equation} 
in the {\em thermodynamic limit} of large $N$ under some statistical assumptions.
For simplicity, we assume \emph{data-model matching}, ie. the probabilistic model \eqref{Gausslate} describes the generation of the dataset $\{\matr y, \matr K\}$ correctly.
Moreover, in order to allow for some nontrivial dependencies 
between matrix elements $K_{ij}$, 
we assume that $\matr K$ is drawn from a rotation invariant matrix ensemble, i.e. $\matr K$ and $\matr V\matr K\matr V^\top$ have the same probability distributions for any orthogonal matrix $\matr V$ independent of $\matr K$. Equivalently, we have the spectral decomposition 
\begin{equation}
\matr K=\matr O^
\top\matr D\matr O \label{decom}
\end{equation}
where $\matr O$ is a Haar (orthogonal) matrix that is independent of a diagonal matrix $\matr D$ \cite{collins2014integration}.  For the perceptron model, the ``classic'' assumption of 
independent components for inputs leads to a {\em Wishart} distribution for 
$\matr K$ which is rotational invariant. For more complex ensembles, see \cite{CakmakOpper19}.
\section{The TAP Equations}
The TAP approach \cite{Mezard},\cite{Adatap}---related to \emph{expectation consistent} inference 
approximations in machine learning \cite{OW5}---typically provides highly accurate approximations for probabilistic inference. In our context, the TAP equations are the fixed-point equations for approximate posterior means $\matr m$ that are given by
\begin{subequations}
\label{TAP}
\begin{align}
\matr m&=m_\nu(\matr \rho,\matr y)\\
\matr \rho&=\nu\matr m-\matr K^{-1}\matr m\\
\chi&=\langle m'_\nu(\matr \rho,\matr y) \rangle\label{chi}\\ 
\nu&={\rm R}(-\chi).
\end{align}
\end{subequations} 
Here, we denote the empirical average of an $N\times 1$ vector $\matr x$ by $\langle\matr x\rangle\doteq \frac 1 N \sum_{i\leq N}x_i$. Moreover, given the auxiliary single-site partition function 
\begin{align}
Z_\nu(\rho,y)\doteq \int {\rm d }\theta\;p(y\vert \theta) e^{-\frac{\nu}{2}\theta^2+\rho\theta}
\end{align}
we introduce the nonlinear functions
\begin{align}
m_\nu(\rho,y)&\doteq \frac{{\partial}\ln{Z_\nu(\rho,y)}}{{\partial}\rho}\\
m_\nu'(\rho,y)&\doteq \frac{{\partial }m_\nu(\rho,y)}{{\partial}\rho}.
\end{align}
The only dependency the TAP equations of the random matrix ensemble is via 
the function ${\rm R}(\cdot)$ which is the so-called R–transform. Its definition will be given later
in equation \eqref{Rtrans}. The TAP equations generalize the naive mean field approximation,
which neglects statistical dependencies between the components of $\matr \theta\sim p(\matr \theta \vert\matr y,\matr K)$. Since the diagonal entries of rotation invariant random matrices
are asymptotically self averaging (e.g. \cite[Theorem 2.1]{cakmak2016random},\cite{Opper16}), a short computation shows that
the naive mean field theory corresponds to the approximation ${\rm R}(-\chi)\approx {\rm R}(0) =
\frac{1}{N}\mbox{tr}(\matr K^{-1})$.  The (Onsager-) correction to the mean naive field 
approximation in terms of the R-transform  was originally derived in \cite{Parisi} 
for Ising spin-glasses using a free energy approach, see also \cite{maillard2019hightemperature} for a comprehensive exposition in this approach. An alternative derivation was given in \cite{Adatap} using the cavity method. We also refer the reader to \cite{CakmakOpper18} for a rigorous approach on the self averaging assumptions made for cavity field variances in the latter approach. 

The function ${\rm R}(\cdot)$ stands for the R–transform of the spectral distribution of $\matr K^{-1}$  which is defined as \cite{mingo2017free}
\begin{equation}
{\rm R}(\omega)\doteq{\rm G}^{-1}(\omega)-\frac{1}{\omega},\quad  -q<\omega<0.  \label{Rtrans}
\end{equation}
Here, we have defined 
\begin{equation}
q\doteq\frac{1}{N}{\rm tr}(\matr K)
\end{equation}
and ${\rm G}^{-1}$ is the functional inverse (which is well-defined on $(-q,0)$) of the Green-function 
\begin{equation}
{\rm G}(z)\doteq \frac{1}{N}{\rm tr}((z{\bf I}-\matr K^{-1})^{-1}) . \label{Greens}
\end{equation}

The TAP method is known to be consistent with the replica-symmetric (RS) ansatz 
when an average over data $\matr y$ and couplings $\matr K$ is considered. The validity
of RS is commonly assumed to be asymptotically exact in the case of \emph{data-model matching} 
\footnote{This conjecture was proven for Wishart distributed $\matr K$ by \cite{barbier2019optimal}}. 
For the case of \emph{data-model matching}, one can show that (see, e.g. the study of \cite{Kab08}), the RS ansatz leads to the asymptotic ($N\to\infty$)
solution for $\chi$ in \eqref{chi} as
\begin{align}
\chi=\mathbb E[m'_{\rm R(-\chi)}(\rho, y)].\label{replica}
\end{align}
Here, the expectation is taken with respect to the probability distribution 
\begin{equation}
p_{rs}(\theta,y,\rho)\doteq N(\theta\vert 0,q)p(y\vert \theta)N(\rho\vert \kappa\theta,\kappa),\quad \kappa\doteq{\rm R}(-\chi)-q^{-1}
\end{equation}
with $N(\cdot\vert \mu,\sigma^2)$ denoting the Gaussian density function with mean $\mu$ and variance $\sigma^2$. This corresponds to the asymptotic marginal  distribution for  the components of $(\matr \theta,\matr y,\matr \rho)$ 
\begin{equation}
(\theta_i,y_i,\rho_i)\sim p_{rs}(\theta_i,y_i,\rho_i). \label{static}
\end{equation}
valid for a large system. Our general idea is to design an iterative algorithm for solving $\matr \rho $ in \eqref{TAP} in terms of an iteration of a vector of auxiliary variables
$\matr \rho(t)$ (for $t=1,\ldots$ denoting the discrete time index of the iteration) such that the ``effective'' distributions of the marginals $(\theta_i,y_i,\rho_i(t))$ mimic the static distribution $p_{rs}(\theta,y,\rho)$, specifically 
\begin{equation}
(\theta_i,y_i,\rho_i(t))\sim N(\theta_i\vert 0,q)p(y_i\vert \theta_i)N(\rho_i(t)\vert \kappa(t)\theta,\sigma^2(t))
\end{equation}
for some dynamical order parameters $\kappa(t)$ and $\sigma^2(t)$ converging both to $\kappa$ as $t\to \infty$. 

\section{The iterative algorithm}
In recent years, there has been considerable interest in analyzing EP-style \cite{Minka1}  iterative algorithms \cite{VAMP,takeuchi2017rigorous,fletcher2018inference}---commonly referred as the method of VAMP. From a computational complexity point of view, such algorithms require the computation of products of $N\times N$ matrices at every iteration steps 
(for details, see \ref{vamp}) which can be problematic for large $N$ and large times. 
Here, we propose a simplified algorithm which utilities the result of the RS ansatz \eqref{replica} and therefore assumes that the assumed data generating process is correct.

We propose the following iterative algorithm for solving the TAP equations~\eqref{TAP}
\begin{subequations}
	\label{alg}
	\begin{align}
	\eta(t)&=\langle m_\nu'(\matr \rho(t-1),\matr y)\rangle\\
	\matr \rho(t)&=\matr Af_{\eta(t)}(\matr \rho(t-1),\matr y).
	\end{align}
\end{subequations}
Here, for short, we have introduced the scalar function
\begin{align}
f_\eta(\rho,y)&\doteq\frac{1}{\eta}m_\nu(\rho,y)-\rho \label{f_t}.
\end{align}
This resembles the update of a single layer recurrent neural network: A vector of nodes
$\matr\rho(t-1)$ is passed point wise through a nonlinear function $f_{\eta(t)}$. The resulting vector
is multiplied by a symmetric ``coupling'' matrix $\matr A$ before the process is repeated. Note that the iterations \eqref{alg} are solely based on matrix vector multiplications and evaluations of scalar nonlinear functions. 

Before iteration starts the iterative algorithm requires the elements $\matr A$ and $\nu$ that are obtained as follows: We first compute the
spectral decomposition
\begin{equation}
\matr K=\matr O^\top {\rm diag}(\matr d)\matr O
\end{equation}
where the vector $\matr d$ contains the eigenvalues of $\matr K$. 
Then, we obtain the parameters $\nu$, $\lambda$ and $\chi$ by solving the fixed--point equations
\begin{subequations}
	\label{vcom}
	\begin{align}
	\chi&=\mathbb E[m'_\nu(\rho, y)] \label{exp}\\
	\lambda&=\frac{1}{\chi}-\nu\\
	\nu &=\left(\frac{1}{N}\sum_{i\leq N}\frac{d_i}{\lambda d_i+1}\right)^{-1}-\lambda.
	\end{align}
\end{subequations}
Finally, the matrix $\matr A$ is computed as
\begin{equation}
\matr A\doteq \frac{1}{\chi}\matr O^\top {\rm diag}(\matr d)(\lambda{\rm diag}(\matr d)+{\bf I})^{-1}\matr O-{\bf I}. \label{Acom}
\end{equation}

It is easy to see that the fixed points of $\matr \rho(t)$ coincide with 
the solution of the TAP equations \eqref{TAP} for $\matr\rho$ given that the empirical average \eqref{chi} is substituted by the RS result \eqref{replica}. In fact, we will re-derive the RS result from the DFT analysis of the iterative algorithm.
\section{The results of DFT}
In this section, we will analyze the dynamical properties of the iterative algorithm \eqref{alg} using the method of the DFT. To this end, we introduce the moment \emph{generating functional} for the trajectory of $\{\rho_i(t)\}\doteq\{\rho_i(t)\}_{t\leq T}$ as
\begin{align}
	Z\{l(t)\}\doteq\int \prod_{t=1}^T & {\rm d}\matr \rho(t)\;  \delta\left[\matr \rho(t)-\matr A{f}_{\eta(t)}(\matr \rho(t-1),\matr y)\right]e^{{\rm i}\rho_i(t) l(t)}\label{mgf}
\end{align}
where for the sake of compactness of notation we assume that the dynamical order parameter $\eta(t)$ is self-averaging. Note, that we have added a single external field $l(t)$ to node $i$.
To obtain statistical averages we compute the averaged-generating functional $\mathbb E[Z(\{l(t)\})]$ where the expectation is taken over the random elements $\matr y, \matr \theta,\matr O$, i.e.
\begin{equation}
	\mathbb E[Z(\{l(t)\})]\doteq\int{\rm d}\matr \theta{\rm d}\matr y{\rm d}\matr O \; Z(\{l(t)\}) {\new p(\matr y,\matr \theta,\matr O\vert \matr d)}.
\end{equation}
	From this, for example, we could compute 
	\begin{align}
	\frac{1}{N}\mathbb E[\matr \rho(t)^\top\matr \rho(s)]&=\mathbb E[\rho_i(t)\rho_i(s)] \label{ex}\\
	&=-\left.\frac{\partial\mathbb E[Z(\{l(t)\})]}{\partial l(t)\partial l(s)}\right|_{\{l(t)\}=0} \label{dfmse}
	\end{align}
	where the identity \eqref{ex} follows from the fact that probability distribution of $\matr A$ is invariant under permutation. Using \eqref{dfmse} we can quantify the averaged-normalized-square Euclidean distance between iterates of the algorithm at different times as
	\begin{equation}
	\frac{1}{N}\mathbb E[\Vert\matr \rho(t)-\matr \rho(s)\Vert^2]=\mathbb E[(\rho_i(t)-\rho_i(s))^2]
	\end{equation}
	which will allow us to compute the convergence properties of the algorithm.
	We will defer the explicit and lengthy computation of DFT to \ref{derDF}.
	There we show how to integrate out the trajectories for all other model $j\neq i$
	in the limit $N\to\infty$. The result is 
	\begin{align}
	\mathbb E[Z(\{l(t)\})]\simeq \int &{\rm d}\mathcal N\left(\{\phi (t)\}\vert \matr 0,\mathcal{{C}}_\phi\right){\rm d}\mathcal N(\theta\vert 0,q){\rm d}y\;p(y\vert \theta)\nonumber \times\\ &\times \prod_{t=1}^{T}{\rm d}\rho(t)\; \delta [\rho(t)-\phi(t)-\kappa(t)\theta]e^{{\rm i}\rho(t)l(t)} \label{rdfg}
	\end{align}
where $\mathcal N(\cdot\vert\matr\mu,\matr\Sigma)$ stands for a Gaussian distribution function with mean $\matr \mu$ and covariance~$\matr \Sigma$. Hence, we have obtained an \emph{effective stochastic process} for the dynamics of single, arbitrary component $\rho(t)$ of the vector $\matr\rho(t)$ as
	\begin{subequations}
		\label{sp}
		\begin{align}
		\{\phi(t)\}&\sim N(\matr 0,\mathcal C_\phi)\\
		(\theta,y)&\sim N(\theta\vert 0,q)p(y\vert \theta)\\
		\rho(t)&=\phi(t)+\kappa(t)\theta.\label{rho}
		\end{align}
	\end{subequations}
The order parameter $\kappa(t)$ and the two-time covariance matrix $\mathcal C_\phi(t,s)$ (which is denoting the $(t,s)$th indexed entries of $\mathcal C_\phi$) of the Gaussian process
are computed by the recursions
	\begin{align}
		\kappa(t)&=\frac{\kappa}{q\lambda}{\mathbb E[\theta \gamma(t)]}\\
	\mathcal C_\phi(t,s)&=\sigma_\matr A^2\mathbb E[\gamma(t)\gamma(s)]+ \frac{\kappa(t)\kappa(s)}{\kappa^2}(\kappa-\sigma^2_{\matr A}(\lambda +q\lambda^2)) \label{covr}.
	\end{align} 
Here, $\sigma_\matr A^2$ stands for the variance of the spectral distribution of $\matr A$ and for short we have introduced the random dynamics
	\begin{equation}
	\gamma(t)\doteq f_{\chi(t)}(\rho(t-1),y) \quad \text{with}\quad  \chi(t)\doteq \mathbb E[m'_{\nu}(\rho(t-1),y)].
	\end{equation} 
The relative simplicity of these equations may come as a surprise to readers familiar with the results of DFT obtained for spin-glass dynamics and neural networks. 
In contrast to models with {\em non symmetric} random matrices (see e.g \cite{sompolinsky1988chaos}) the symmetric matrix case
is usually plagued with memory terms in the effective single node stochastic processes. These usually make explicit analytical computations of single time marginals a hard task. As shown in \ref{gmfc} the absence of such memory terms can be explained by the vanishing of the two-time response functions which in turn can be understood by well-known concepts of random matrix theory.

\subsection{Convergence rate of the algorithm}
To study the large time behavior of the single node dynamics we define the deviation between the dynamical variables at different times ($t\neq s$) as
	\begin{align}
	\Delta_\rho(t,s)&\doteq \mathbb E[(\rho(t)-\rho(s))^2]\\
	&=\mathcal C_\rho(t,t)+\mathcal C_\rho(s,s)-2\mathcal C_\rho(t,s)\label{decdelta}
	\end{align}
where $\mathcal C_\rho(t,s)$ stands for the two-time covariance of the zero-mean Gaussian process $\{\rho(t)\}$. We have defined the rate of convergence as 
\begin{equation}
\mu_\rho\doteq\lim_{t,s\to \infty} \frac{\Delta_\rho(t+1,s+1)}{\Delta_\rho(t,s)}.
\end{equation}
Assuming convergence, that is $\Delta(t,s)\to 0$ as $t,s\to \infty$, we get the explicit rate as 
	\begin{align}
	\mu_\rho= \frac{\sigma_\matr A^2}{\chi^2}(\mathbb E[m'_\nu(\rho,y)^2]-\chi^2). \label{rc}
	\end{align}
{\new Furthermore, it turns out that $\ln\mu_{\rho}$ gives the exponential decay rate as
\begin{equation}
\lim_{t\to \infty}\frac{1}{t}\ln \Delta_\rho(t,\infty)=\ln \mu_{\rho}\label{newrate}
\end{equation}
with~$\Delta_\rho(t,\infty)\doteq \lim_{s\to\infty}\Delta_\rho(t,s)$. The proofs of \eqref{rc}~and~\eqref{newrate} are given in \ref{drc}.}
	
The spectral variance $\sigma_\matr A^2$ can be expressed in terms of the derivative of the R-transform as (see \cite[Eq.(38)]{CakmakOpper19})
\begin{equation}
\sigma_\matr A^2=\frac{\chi^2{\rm R}'(-\chi)}{1-\chi^2{\rm R}'(-\chi)}.
\end{equation}
Then, we re-write \eqref{rc} in the form
\begin{equation}
\mu_\rho=1-\frac{1-\mathbb E[m'_\nu(\rho,y)^2]{\rm R}'(-\chi)}{1-\chi^2{\rm R}'(-\chi)}.
\end{equation}
Thus, the necessary condition for convergence $\mu_\rho<1$ holds if and only if
\begin{equation}
\mathbb E[m'_\nu(\rho,y)^2]{\rm R}'(-\chi)<1. \label{AT}
\end{equation}
Following the arguments for \cite[Eq.(18)]{takeda2007statistical} one can conclude that equation \eqref{AT} coincides with the stability condition of the RS ansatz - known as  \emph{de Almeida Thouless} (AT) criterion. Since 
for the case of data-model matching, RS solution is usually locally stable, we can expect that local convergence broadly fulfilled. 

\subsection{Asymptotic consistency with the RS ansatz}
Let $\gamma\doteq f_{\chi}(\rho,y)$ where $(\theta, y,\rho)\sim p_{rs}(\theta,y,\rho)$. Then, we have
\begin{align}
\mathbb E[\theta\gamma]&=q\lambda\\
\mathbb E[\gamma^2]&=\lambda +q\lambda^2.
\end{align}	
Hence, if $\kappa(t)$ and $\mathcal C_\phi(t,t)$ converge, they both converge to $\kappa$. Thus, the stationary distribution of the stochastic process \eqref{sp} yields the replica-symmetric solution, i.e.
\begin{equation}
\lim_{t\to\infty}p(\theta,y,\rho(t))=p_{rs}(\theta,y,\rho).
\end{equation}
\section{Simulation results}\label{Sim}
Perceptrons are single layer neural networks that are parameterized by a vector of weights, say $\matr w\in \RR^{K\times 1}$. We consider the binary classification problems from a training set that is given by $\{(\matr x_i ,y_i)\}_{i\leq N}$. Here $\matr x_i\in \RR^{K\times 1}$ stands for a vector of inputs and a binary label $y_i=\mp1$ for classification. Specifically, we have the observation model for $\matr y$ as
\begin{equation}
\matr y={\rm sign}(\matr X\matr w+\matr \epsilon), \quad \matr X\doteq[\matr x_1,\cdots,\matr x_N]^\top.
\end{equation}
The Gaussian latent vector $\matr w\sim N(\matr 0,{\bf I})$ and the noise vector $\matr \epsilon\sim N(\matr 0,\sigma_0^2{\bf I})$ are generated independently. So that, we set $\matr \theta=\matr X\matr \omega$, $\matr K=\matr X\matr X^\top$ and $p(y\vert \theta)=\Theta(y\theta/\sigma_0)$ with $\Theta(\cdot)$ denoting the cumulative distribution function of standard normal distribution. 

We illustrate our theoretical results through the following random matrix models for the data matrix:
\begin{itemize}
	\item [(i)] The entries of $\matr X$ are independent Gaussian with zero mean variance $1/N$;
	\item [(ii)] $\matr X=\tilde{\matr H}\matr P$ where $\matr P$ is the $N\times K$ projection matrix with $N\geq K$ and $\matr P_{ij}=\delta_{ij}$, and $\tilde{\matr H}$ is an $N\times N$ a \emph{randomly-signed} Hadamard matrix \cite{Greg} as
	\begin{equation}
	\tilde {\matr H}=\frac{1}{\sqrt{N}}\matr Z\matr H_N.
	\end{equation}
	Here $\matr Z$ is a uniformly distributed random $N\times N$ signed permutation matrix, specifically $Z_{ij}=\epsilon_i\delta_{i,\sigma(j)}$ for $\epsilon_i= \mp1$
	being independent binary random variables with equal probabilities and $\sigma$ is a random permutation. Furthermore, $\matr H_N$ is the $N\times N$ Hadamard matrix which is {\em deterministically} constructed
	from the recursion 
	\begin{align}
	\matr H_{2^k}=\left[\begin{array}{cc}
	\matr H_{2^{k-1}}& \matr H_{2^{k-1}}\\
	\matr H_{2^{k-1}}&-\matr H_{2^{k-1}}
	\end{array}\right] ~~ \text{with}~~ \matr H_{1}\doteq 1.
	\end{align}
\end{itemize}
Note, that in the latter model, $\matr X$ is a column-orthogonal matrix (i.e. $\matr X^\top\matr X={\bf I}$) with the {\em binary entries} $X_{ij}=\mp \frac{1}{\sqrt{N}}$. Though, $\matr K=\matr X\matr X^\top$ is not rotation invariant in this case,  motivated with the study \cite{Greg} we expect that our theoretical results might still yield quite accurate approximations. 

\begin{figure}[h]
	\epsfig{file=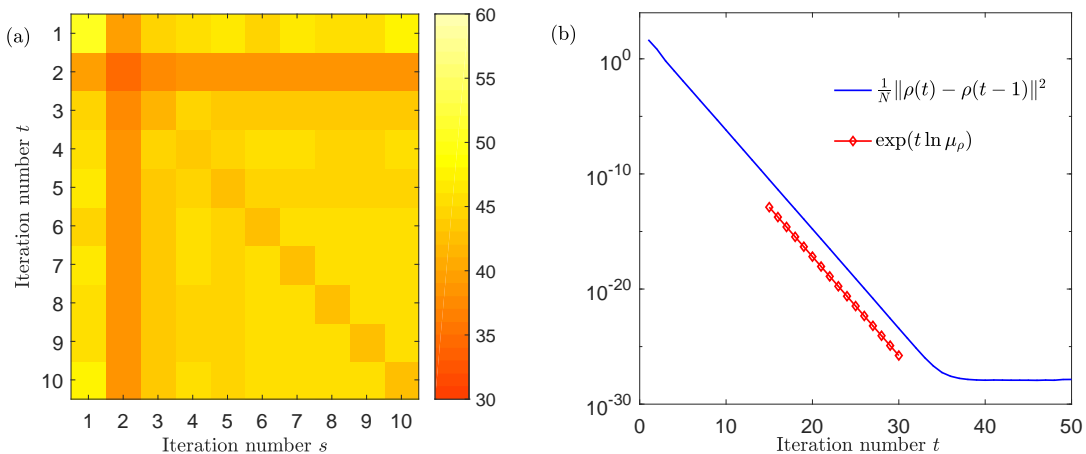,width=1.06\columnwidth}
	\caption{Random matrix model (i):  The model parameters are chosen as $\sigma_0^2=10^{-2}$, $N=2K$ and $K=10^{4}$. (a) Discrepancy between theory and simulations for the two-time covariance with the $(t,s)$ indexed segment representing the relative-squared-error in dB, i.e. $-10\log_{10}{rse}(t,s)$. (b) Asymptotic of the algorithm (where the flat line around $10^{-30}$ are the consequence of the machine precision of the computer which was used).}\label{fig1}
\end{figure}

\begin{figure}[h]
	\epsfig{file=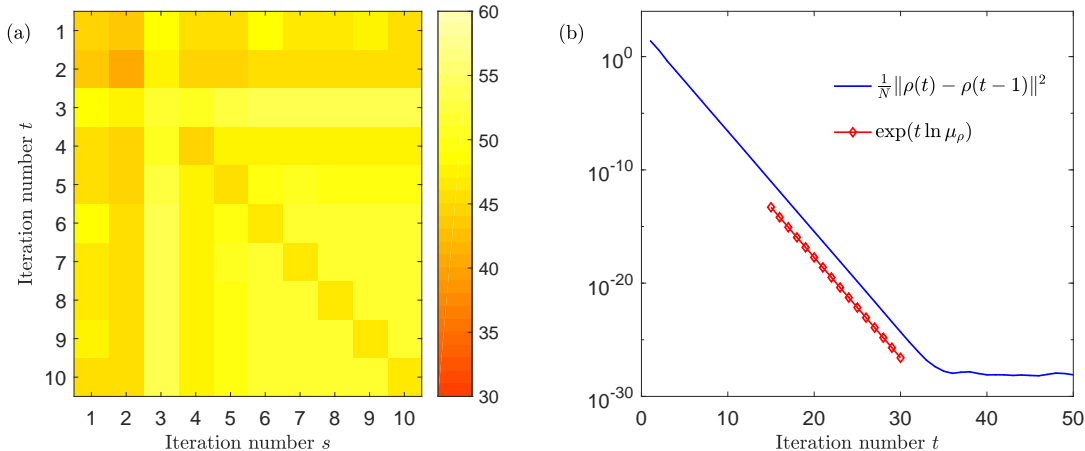,width=1.06\columnwidth}
	\caption{Random matrix model (ii): The model parameters are chosen as $\sigma_0^2=10^{-2}$, $N=2K$ and $K=2^{13}$. (a) Discrepancy between theory and simulations for the two-time field covariance with $(t,s)$ indexed segment representing  the relative-squared-error in dB, i.e. $-10\log_{10}{\rm rse}(t,s)$. (b) Asymptotic of the algorithm.}\label{fig2}
\end{figure}

Figure~\ref{fig1} refers to random matrix model (i) where in Figure\ref{fig1}-(a) and (b) we illustrate the discrepancy between theory and simulations for the two-time covariance $\mathcal C_\rho(t,s)$ with respect to the two-time relative-square-error 
\begin{equation}
{rse}(t,s)\doteq\left(\frac{\mathcal C_\rho(t,s)-\frac{1}{N}\matr \rho(t)^\top \matr \rho(s)}{\mathcal C_\rho(t,s)}\right)^2
\end{equation}
and the analytical convergence rate of the algorithm, respectively. Similarly, Figure~\ref{fig2} refers to the random matrix model (ii). Simulation results are based on \emph{single instances} of large random matrices and data. 
\section{Summary and Outlook}
We have presented the analysis of a Bayesian inference algorithm for latent Gaussian models in the large systems limit. We have based our approach on a statistical mechanics path integral technique, dynamical functional theory (DFT), which has been used before to treat spin-glass models  and neural networks with random interactions. We have generalized the method to
allow for more complex settings  of the quenched randomness, allowing for a combination
of a teacher-student scenario together with rotation invariant ensembles of coupling matrices.
While related inference algorithms had been treated before using rigorous random matrix
techniques, we were able, for the first time, to obtain the complete effective marginal
stochastic process of single nodes. Although the computations turned out to be somewhat
involved, the final results turned out to be surprisingly simple: The statistics of the
effective field entering the nonlinear function of the algorithm was found to be a Gaussian
process, for which the covariance function could be obtained iteratively. This result lacked
the complications of the non-Gaussian fields caused by memory terms in the effective dynamics which were typically present in spin-glass dynamics.
We could trace the origins of this simplification by utilizing concepts of random matrix   
theory. Based on our general result, we were able to compute exact convergence
rates of the algorithm. The statistics of the algorithm's fixed points coincides with that
predicted by the replica theory. Within our scenario of data-model matching, the algorithm is
found (at least) to be locally convergent.  Simulations on single, large systems showed excellent agreement with the theory, From a more practical point of view,
the restriction to the matching case allowed us to define an inference algorithm in terms of more
efficient iterations (compared to previous VAMP algorithms) by using only a single fixed matrix.

We expect that our approach can be extended in various directions. The DFT method is
general enough to treat the non-matching case as well, including the settings of VAMP algorithms.
It will also be interesting to extend the method to other more complex probabilistic models of
neural network type such as the {\em restricted Boltzmann machines} \cite{Tramel18}. 
This however, would require different types of integrals over random matrices,
which have to be incorporated into the DFT formalism. A further interesting generalization of
DFT would be to random sequential updates in algorithms, where at each iteration step
only a single, randomly selected node (or a small mini batch of nodes) are used in the update.
It will be interesting to see, if a properly adapted DFT method would lead to tractable
computations for this more realistic scenario.

The statistical mechanics techniques utilized in our DFT approach might present of course
certain limitations of applicability to real world machine learning problems. The study
of general rotation invariant ensembles of matrices is a nontrivial improvement
over older models that were based on simpler independence assumptions.
It  allows us to deal with matrices of (almost) arbitrary eigenvalue distributions, but restricts
the corresponding eigenvectors as irrelevant, simply pointing in random directions.
It will important to find out for which types of real data such technical assumptions might be reasonable
enough to draw relevant conclusions from the theory. The excellent agreement
between theory and simulations for the case of a randomly--signed Hadamard matrix
(which contains much less randomness compared to a the rotation invariant case, see section \ref{Sim})
gives us some hope that our method could be applicable to a broader range of problems.
\section*{Acknowledgment}
This work was supported by the German Research Foundation, Deutsche Forschungsgemeinschaft (DFG), under Grant ``RAMABIM'' with No. OP 45/9-1.
\appendix
\section{The VAMP algorithm and a generic memory-free construction}
\subsection{VAMP algorithm}\label{vamp}
In our context, the VAMP algorithm \cite{VAMP,takeuchi2017rigorous} coincides with the iterative process
\begin{subequations}
\label{VAMP}
	\begin{align}
	\eta(t)&=\langle m_{\nu(t-1)}'(\matr \rho(t-1),\matr y)\rangle\\
	\lambda(t)&=\frac{1}{\eta(t)}-\nu(t-1)\\
	\tau(t)&=\frac{1}{N}\sum_{i\leq N} \frac{d_i}{\lambda(t) d_i+1}\\
	\nu(t)&=\frac{1}{\tau(t)}-\lambda(t)\\
	\matr A(t)&=\frac{1}{\tau(t)}\matr O^\top {\rm diag}(\matr d)(\lambda(t){\rm diag}(\matr d)+{\bf I})^{-1}\matr O-{\bf I}\label{At}\\
	\matr \rho(t)&=\matr A(t) \left[\frac{1}{\eta(t)}m_{\nu(t-1)}(\matr \rho(t-1),\matr y)-\matr\rho(t-1)\right].
	\end{align}	
\end{subequations}
From the computational complexity point of view, the essential difference between our proposed iterations \eqref{alg} from the VAMP iterations \eqref{VAMP} is that at every iteration step the former requires a matrix-vector multiplication while the latter requires matrix-matrix multiplications (see \eqref{At}).
\subsection{A generic memory-free dynamical construction}\label{gmfc}
The striking property of the algorithm \eqref{alg} is that it has the memory-free property
\begin{equation}
\mathbb E\left[\frac{\partial\rho_i(t)}{\partial\rho_i(s)}\right]\simeq 0 \quad \forall t,s.
\end{equation}
This property makes the analysis of the algorithm relatively simple. To have a better insight on designing memory-free algorithms, we will next introduce a generic dynamical construction and present an intuitive  derivation of its memory-free property by means of the concept of asymptotic freeness of random matrices \cite{Hiai,mingo2017free}. 

Specifically, consider the following generic iterative process
\begin{equation}
	\matr \rho(t)=\matr A(t)f_{t}(\matr \rho(t-1))
\end{equation}
where $f_{t}$ is a sequence of scalar function. Here, we suppose that $\matr A(t)$ is rotation invariant (for all $t$)
and also for the diagonal matrices 
\begin{equation}
[\matr E(t)]_{ij}\doteq\frac{\partial f_{t}( \rho_i(t-1))}{\partial\rho_i(t-1))}\delta_{ij}
\end{equation}
we suppose that 
\begin{align}
	{\rm Tr}(\matr A(t))=0 \quad \text{and} \quad {\rm Tr}(\matr E(t))=0.  \label{centered}
\end{align}
Here, for an $N\times N$ matrix $\matr X$ we denote its limiting normalized-trace by 
\begin{equation}
{\rm Tr}(\matr X)\doteq \lim_{N\to\infty}\frac{1}{N}{\rm tr}(\matr X).
\end{equation}
Remark that both our proposed algorithm \eqref{alg} and the VAMP algorithm \eqref{VAMP} are in the family of this generic dynamical construction.   

We are interested the analyzing the diagonal elements of the dynamical susceptibility
\begin{equation}
\frac{\partial{\rho_i}(t)}{\partial\rho_j(s)}=
\left[({\matr A}(s+1)\matr E(s+1){\matr A}(s+2)\matr E(s+2)\cdots{\matr A}(t) {\matr E(t)})\right]_{ij} \label{jacob}.
\end{equation}
In the large-system limit, this product of matrices can be simplified by
the concept of \emph{asymptotic freeness} of random matrices. Specifically, for the two families of matrices, say ${\mathcal A\doteq \{\matr A_1,\matr A_2,\ldots,\matr A_a\}}$ and ${\mathcal E\doteq \{\matr E_1,\matr E_2,\ldots,\matr E_e\}}$, let ${\matr P_i(\mathcal A)}$ and ${\matr Q_i(\mathcal E)}$ stand for (non-commutative) polynomials of the matrices in ${\mathcal A}$ and the matrices in ${\mathcal E}$, respectively. Then,  we say the families ${\mathcal A}$ and ${\mathcal E}$ are asymptotically free if for all $i\in[1,K]$
and for all polynomials ${\matr P_i(\mathcal A)}$ and ${\matr Q_i(\mathcal E)}$ we have \cite{Hiai}
\begin{equation*}
{\rm Tr}({\matr P_1(\mathcal A)}{\matr Q_1(\mathcal E)}{\matr P_2(\mathcal Q)}{\matr Q_2(\mathcal E)}\cdots {\matr P_K(\mathcal A)}{\matr Q_K(\mathcal E)})=0
\end{equation*}
given that 
\begin{equation*}
{\rm Tr}({\matr P_i(\mathcal A)})={\rm Tr}({\matr Q_i(\mathcal E)})=0, \quad \forall i.
\end{equation*}
In other words, the limiting normalized-trace of any adjacent product of powers of matrices -- which belong to different free families and are centered around their limiting normalized-traces -- vanishes asymptotically.

The matrices in \eqref{jacob} belong to two families: rotation invariant and diagonal. Under certain technical conditions, these two matrix families can be treated as asymptotically free (in the almost sure sense) \cite{Hiai}, i.e. \emph{any adjacent two matrices in the susceptibility matrix \eqref{jacob}~are asymptotically free}. Moreover, by design the matrices are \emph{centered around their limiting normalized-traces} \eqref{centered}. Hence, it immediately follows from the definition of asymptotic freeness that
\begin{equation}
\lim_{N\to\infty}\frac{1}{N}\sum_{i\leq N}\frac{\partial\rho_i(t)}{\partial\rho_i(s)}=0. \label{v1}
\end{equation}	
In fact, following the analysis of \cite{CakmakOpper18} we can obtain a stronger result
\begin{equation}
\lim_{N\to\infty}\frac{1}{N}\sum_{i\leq N}\left(\frac{\partial\rho_i(t)}{\partial\rho_i(s)}\right)^2=0.
\end{equation}
This implies the self-averaging property
\begin{equation}
\left\{\frac{\partial\rho_i(t)}{\partial\rho_i(s)}\right\}_{i\leq N}{\small \to} 0.
\end{equation}

\section{Derivation of the results of dynamical functional}\label{derDF}
Our first goal is to perform the average
\begin{align}
\mathbb E[Z(\{l(t)\})]=\int{\rm d}\matr \theta{\rm d}\matr y{\rm dP}(\matr O)\; Z(\{l(t)\}) p(\matr y\vert  \matr \theta )N(\matr \theta\vert\matr 0,\matr K)
\end{align}
where ${\rm dP}(\matr O)$ stands for the Haar invariant measure of the orthogonal group $O(N)$. To this end, we first invoke the representations of the Dirac $\delta$ functions in terms of its characteristic function and write the generating functional \eqref{mgf} of the form
\begin{align}
Z(\{l(t)\})=&\int \prod_{t\leq T} {\rm d}\matr\rho(t){\rm d}{\matr\gamma}(t)\;\delta\left[\matr{\gamma}(t)-{f}_{\eta (t)}(\matr \rho(t-1),\matr y)\right]\delta\left[\matr \rho(t)-\matr A\matr \gamma(t)\right]e^{{\rm i} \rho_i(t)l(t)} \\
=&c\int \prod_{t\leq T} {\rm d}\matr \rho(t){\rm d}{\matr\gamma}(t){\rm d}\hat{\matr \rho}(t)\; \delta\left[\matr{\gamma}(t)-{f}_{\eta (t)}(\matr \rho(t-1),\matr y)\right]e^{{\rm i}\hat{\matr \rho}(t)^\top[{\matr \rho}(t)-\matr A\matr \gamma(t)]}e^{{\rm i} \rho_i(t)l(t)} \label{com1}
\end{align} 
Here, and throughout the sequel, $c$ stands for a constant term---which will irrelevant for the analysis--- to ensure the normalization property $\mathbb E[Z(\{l(t)=0\})]=1$.
Moreover, we note, from the representation of the Gaussian density function in terms of the characteristic function, 
that 
\begin{align}
N(\matr \theta\vert\matr 0,\matr K)&=c e^{\frac{\lambda}{2}\matr \theta^\top\matr \theta}\int {\rm d}\matr u\;e^{\frac{{\rm i}}{\sqrt{\chi}}\matr u^\top\matr\theta} e^{-\frac{1}{2} \matr u^\top \frac{1}{\chi}\matr K(\lambda\matr K+{\bf I})^{-1}\matr u}\\
&=ce^{\frac{\lambda}{2}\matr \theta^\top\matr \theta}\int {\rm d}\matr u \; e^{\frac{{\rm i}}{\sqrt{\chi}}\matr u^\top\matr\theta}e^{-\frac{1}{2}\matr u^\top\matr u} e^{-\frac{1}{2} \matr u^\top\matr A\matr u}.\label{com2}
\end{align}
Then, by invoking \eqref{com1} and \eqref{com2} we write the averaged-generating functional as 
\begin{align}
\mathbb E[Z(\{l(t)\})]=&c\int {\rm d}\matr \theta {\rm d}\matr y {\rm d}\matr u\; p(\matr y\vert\matr \theta)e^{\frac{\lambda}{2}\matr\theta^\top\matr \theta}e^{-\frac{1}{2}\matr u^\top\matr u}e^{\frac{{\rm i}}{\sqrt{\chi}}\matr u^\top\matr \theta}\; \times \nonumber \\&~~~~\times\prod_{t\leq T} {\rm d}\matr \rho(t){\rm d}{\matr\gamma}(t){\rm d}\hat{\matr \rho}(t) \delta\left[\matr{\gamma}(t)-{f}_{\eta (t)}(\matr \rho(t-1),\matr y)\right]e^{{\rm i}\hat{\matr \rho}(t)^\top{\matr \rho}(t)}e^{{\rm i}\nonumber \rho_i(t)l(t)}\nonumber \\ & ~~~~\times \mathbb E_{\matr O}\left[e^{-\frac{1}{2}\matr u^\top \matr A\matr u -{\rm i}\sum_{t\leq T}\hat{\matr \rho}(t)^\top\matr A{\matr \gamma}(t)}  \right].\label{ag}
\end{align}
\subsection{Disorder average}
In this subsection we will perform the disorder average, i.e. the last term of \eqref{ag}. To this end,  we introduce  the $N\times 1$ vector $\matr z\doteq \frac{\matr u}{\sqrt{N}}$ and the $N \times T$ matrices ${\matr {X}}$ and $\matr {\hat X}$ with the entries $X_{it}\doteq\frac{\gamma_i(t)}{\sqrt{N}}$ and $\hat X_{it}\doteq \frac{\hat \rho_i(t)}{{\rm i}\sqrt{N}}$. Moreover, let 
\begin{equation}
{\matr Q}\doteq\matr{\hat X} \matr {X}^\top+\matr {X}\matr {\hat X}^\top.
\end{equation}
So that we write 
\begin{align}
-\frac{1}{2}\matr u^\top \matr A\matr u=-\frac{N}{2}{\rm tr}(\matr A\matr z\matr z^\top)\quad \text{and}\quad -{\rm i}\sum_{t\leq T}\hat{\matr \rho}(t)^\top\matr A{\matr \gamma}(t)=\frac{N}{2}{\rm tr}(\matr A\matr Q).
\end{align}
Thus, we have
\begin{align}
\mathbb E_{\matr O}\left[e^{-\frac{1}{2}\matr u^\top \matr A\matr u -{\rm i}\sum_{t\leq T}\hat{\matr \rho}(t)^\top\matr A{\matr \gamma}(t)}  \right]&=\mathbb E_{\matr O}\left [e^{\frac{N}{2}{\rm tr}(\matr A({\matr Q}-\matr z\matr z^\top))}\right]. \label{doa}
\end{align}
Using the asymptotic Itzykson-Zuber integral formulation \cite{Collins5,Guionnet} the expectation \eqref{doa} can be expressed in terms of the so-called free cumulants of the spectral distribution of $\matr A$ as
\begin{equation}
\mathbb E_{\matr O}\left [e^{\frac{N}{2}{\rm tr}(\matr A({\matr Q}-\matr z\matr z^\top))}\right]=e^{\frac N 2 (\epsilon_N+\sum_{n=1}^{\infty}\frac{c_{\matr A,n}}{n}{\rm tr}(({\matr Q}-\matr z\matr z^\top)^n))}
\end{equation}
where $c_{\matr A,n}$ stands for the $n$th order free cumulant of the spectral distribution of $\matr A$ and the constant term $\epsilon_N\to0$ as $N\to \infty$. In particular, $c_{\matr A,1}$ and $c_{\matr A,2}$ are the mean and variance of the distribution, respectively, i.e. 
\begin{equation}
c_{\matr A,1}=\frac{1}{N}{\rm tr}(\matr A)=0\quad \text{and}\quad c_{\matr A,2}=\sigma_\matr A^2. 
\end{equation} 
In fact, the R-transform can be defined as a generating function of the free cumulants \cite{mingo2017free}
\begin{equation}
{\rm R}_{\matr A}(\omega)=\sum_{n=1}^{\infty}c_{n,\matr A}\omega^{n-1}.\label{RA}
\end{equation}

We will evaluate ${\rm tr}(({\matr Q}-\matr z\matr z^\top)^n)$ in terms of the order parameters 
\begin{align}
\mathcal {G}&\doteq\matr{X}^\top \matr {\hat X} \\
\mathcal{C}&\doteq\matr{X}^\top \matr {X} \\
\mathcal{\tilde{C}}&\doteq\matr{\hat X}^\top \matr {\hat X} \\
\mathcal {B}&\doteq \matr z^\top\matr {X}\\
\mathcal {\tilde B}&\doteq \matr z^\top \hat{\matr X}\\
\zeta&\doteq \matr z^\top\matr z.
\end{align}
Note that ${\matr Q}=[\matr {\hat X}~~\matr {X }] [\matr {X}~~\matr {\hat X}]^\top$. Hence, we have 
\begin{equation}
{\matr Q}^n= [\matr {\hat X}~~\matr{X}]\mathcal Q^{n-1} [\matr {X}~~\matr {\hat X}]^\top\quad \text{with}\quad \mathcal Q\doteq \left[\begin{array}{cc}
\mathcal G & \mathcal {C}\\
\mathcal{\tilde C}&\mathcal G^{\top}
\end{array}\right].
\end{equation}
By the cyclic invariance property of the trace operator,  the traces ${\rm tr}(({\matr Q}-\matr z\matr z^\top)^n)$ do solely depend on the terms $\{\zeta^k\}$ and $\{\matr z^\top\matr Q^m\matr z=  [\mathcal{\tilde B}~~\mathcal {B}]\mathcal Q^{m-1} [\mathcal { B}~~\mathcal {\tilde B}]^\top\}$. Hence, we can define 
\begin{equation}
f_n(\mathcal G,\mathcal {C},\mathcal {\tilde C},\mathcal {B},\mathcal {\tilde B},\zeta)\doteq{\rm tr}(({\matr Q}-\matr z\matr z^\top)^n).
\end{equation}
In particular, we have (see~\ref{sparg})
\begin{align}
f_n(\mathcal G,\mathcal {C},\mathcal {\tilde C},\mathcal {B},\mathcal {\tilde B},\zeta)=&2{\rm tr}(\mathcal G^n)+(-\zeta)^n+ n\sum_{k=0}^{n-2} {\rm tr}\left(\mathcal G^k \mathcal {C} (\mathcal G^\top)^{n-2-k} \mathcal {\tilde C}\right)-2n\sum_{k=0}^{n-2}(-\zeta)^{k}\mathcal{\tilde B}\mathcal G^{n-k-2}\mathcal{B}^\top\nonumber\\
&-n(1-\delta_{n2})\sum_{k=0}^{n-2}(-\zeta)^{k}\sum_{l=0}^{n-k-3}\mathcal{B}\mathcal G^l\mathcal {\tilde C}(\mathcal G^\top)^{n-k-l-3}\mathcal {B}^\top+{\rm SP}(\mathcal G,\mathcal {C},\mathcal {\tilde C},\mathcal {B},\mathcal {\tilde B},\zeta) \label{verygood}
\end{align}
where for $\mathcal{X}\in\{\mathcal {\tilde C},\mathcal {\tilde B}\}$ we have for the last term
\begin{equation}
\left.\frac{\partial {\rm SP}(\mathcal G,\mathcal {C},\mathcal {\tilde C},\mathcal { B},\mathcal {\tilde B},\zeta)}{\partial\mathcal{X}}\right\vert_{\mathcal {\tilde C}=\matr 0, \mathcal {\tilde B}=\matr 0}=\matr 0.
\end{equation}
As usual \cite{Eisfeller,Opper16}, this means that at the saddle-point values $\mathcal {\tilde C}=\matr 0$ and $\mathcal {\tilde B}=\matr 0$ the term ${\rm SP}(\mathcal G,\mathcal {C},\mathcal {\tilde C},\mathcal { B},\mathcal {\tilde B},\zeta)$ does not contribute to saddle--point equations.
\subsection{Saddle-point analysis}
We introduce the single-site (effective, more specifically) generating-functional
\begin{align}
Z_{ef}(\{l(t)\},\mathcal {\hat G},\mathcal {\hat{{C}}},\mathcal {\hat {{\tilde C}}},\mathcal {\hat {B}},\mathcal {\hat {\tilde B}},\hat \zeta)\doteq c\int &{\rm d}\theta{\rm d}y{\rm d}u\; p(y\vert \theta)e^{\frac{\lambda}{2}\theta^2}e^{-\frac{1}{2}u^2}e^{\frac{{\rm i}}{\sqrt{\chi}} u\theta}\prod_{t=1}^{T}{\rm d}{\rho}(t){\rm d}\gamma(t)\hat{\rho}(t)\times
\nonumber\\&\times\delta [{\gamma}(t)- f_{\chi(t)}(\rho(t-1),y)] e^{{\rm i}\hat{ \rho}(t){\rho}(t)}e^{{\rm i}\nonumber \rho(t)l(t)}\\&\times e^{-\sum_{(t,s)}[{\rm i}\mathcal{\hat G}(t,s)\gamma(t){\hat\rho}(s)+{\rm i}\mathcal {\hat { C}}(t,s)\gamma(t){\gamma}(s)+\mathcal{\hat{\tilde C}}(t,s){\hat\rho}(t) {\hat\rho}(s)]}\nonumber \\&\times e^{-{\rm i}\hat{\zeta}u^2-{\rm i}u\sum_{t}[\mathcal {\hat {B}}(t)\gamma(t)+\mathcal {\hat {\tilde B}}(t)\hat\rho(t)]}.
\end{align}
Here, $\chi(t)\doteq \mathbb E[m'_\nu(\rho(t-1),y)]_{Z_{ef}}$ where $\mathbb E[(\cdot)]_{Z_{ef}}$ represents the expectation of the argument with respect to the effective generating-functional $Z_{ef}$.
Then, we write
\begin{align}
\mathbb E[Z(\{l(t)\})]= c\int &{\rm d} \mathcal G {\rm d}\mathcal {\hat G} {\rm d} \mathcal{C} {\rm d}\mathcal{\hat {C}} {\rm d} \mathcal{\tilde C} {\rm d}\mathcal{\hat {\tilde C}}{\rm d}\mathcal{{B}}{\rm d}\mathcal{\hat {B}}{\rm d}\mathcal{{\tilde B}}{\rm d}\mathcal{\hat {\tilde B}}{\rm d}\zeta{\rm d}\hat\zeta\; Z_{ef}(\{l(t)\},\mathcal {\hat G},\mathcal {\hat{{C}}},\mathcal {\hat {{\tilde C}}},\mathcal {\hat {B}},\mathcal {\hat {\tilde B}},\hat \zeta) \nonumber \\
&\times e^{\frac{N}{2}(\epsilon_N+\sum_{n=1}^{\infty}\frac{c_{\matr A,n}}{n}f_n(\mathcal G,\mathcal {C},\mathcal {\tilde C},\mathcal {B},\mathcal {\tilde B},\zeta))}\nonumber \\
&\times e^{N\sum_{(t,s)}[-\mathcal{\hat G}(t,s) \mathcal G(t,s)+{\rm i}\mathcal{\hat {C}}(t,s)\mathcal {C}(t,s) -\mathcal{\hat{\tilde C}}(t,s) \mathcal{\tilde C}(t,s)]}\nonumber\\
&\times e^{N\left({\rm i}\hat \zeta \zeta+\sum_{t}[{\rm i}\mathcal{\hat {B}}(t)\mathcal {B}(t) -\mathcal {\hat{\tilde B}}(t)\mathcal{\tilde B}(t)]\right)}.
\end{align}

In the large $N$ limit, we can perform the integration over $\mathcal G,\mathcal{\hat G},\mathcal { C},\mathcal{\hat {C}},\mathcal{\tilde C},\mathcal{\hat {\tilde C}},\mathcal{{ B}},\mathcal{\hat {B}},\mathcal{{\tilde B}},\mathcal{\hat {\tilde B}},\zeta,\hat \zeta$ with the saddle point method. Doing so yields: 
\begin{align}
\mathcal G(t,s)&=\mathbb E[{-\rm i}\gamma(t)\hat\rho(s)]_{Z_{ef}}\\
\mathcal {C}(t,s)&=\mathbb E[\gamma(t)\gamma(s)]_{Z_{ef}}\\
\mathcal{\tilde C}(t,s)&=\mathbb E[-\hat\rho(t)\hat\rho(s)]_{Z_{ef}} \label{unpsy}\\
\mathcal {B}(t)&=\mathbb E[u \gamma(t)]_{Z_{ef}}\\
\mathcal{\tilde B}(t)&=\mathbb E[-{\rm i}u\hat\rho(t)]_{Z_{ef}}\\
\zeta&=\mathbb E[u^2]_{Z_{ef}}.\label{goodgood}
\end{align}
Furthermore, the solutions $\mathcal{\tilde C}=\matr 0$ and $\mathcal{\tilde B}=\matr 0$  yield from \eqref{verygood} that $\mathcal {\hat {C}}=\matr 0$ and $\mathcal {\hat {B}}=\matr 0$, respectively. Moreover, we have
\begin{align}
\mathcal {\hat G}&={\rm R}_{\matr A}(\mathcal G)\label{Ghat}\\
\mathcal{\hat{\tilde C}}&=\frac{1}{2}\sum_{n=2}^\infty c_{\matr A,n}\sum_{k=0}^{n-2}\mathcal G^k \mathcal {C}(\mathcal G^\top)^{n-2-k}\nonumber \\
&-\frac{1}{2}\sum_{n=3}^\infty c_{\matr A,n}\sum_{k=0}^{n-2}(-\zeta)^{k}\sum_{l=0}^{n-k-3}(\mathcal G^\top)^l\mathcal{ B}^{\top}\mathcal{B}\mathcal G^{n-k-l-3}\label{cov}\\
\mathcal{\hat{\tilde B}}&=-\mathcal{B}\sum_{n=2}^\infty c_{\matr A,n}\sum_{k=0}^{n-2}(-\zeta)^{k}\mathcal G^{n-k-2}\label{Bhat}\\
{\rm i}\hat{\zeta}&=\frac{1}{2}{\rm R}_{\matr A}(-\zeta)
\end{align}
where ${\rm R}_{\matr A}$ stands for the R-transform of the spectral distribution of $\matr A$, see \eqref{RA}. In these equations, we drop the contributions $\frac{\partial \epsilon_N}{\partial \mathcal X}$ for $\mathcal  X=\{\mathcal G,\mathcal {\tilde C},\mathcal {\tilde B},\zeta\}$ at the saddle point analysis, given that $\epsilon_N\simeq 0$. 

In summary, we have $\mathbb E[Z(\{l(t)\})]\simeq Z_{ef}(\{l(t)\})$ where we define
\begin{align} 
Z_{ef}(\{l(t)\})\doteq c\int &{\rm d}\theta{\rm d}y{\rm d}u\; p(y\vert \theta)e^{\frac{\lambda}{2}\theta^2}e^{-\frac{1}{2}\tilde{\rm R}_\matr A  u^2}e^{\frac{{\rm i}}{\sqrt{\chi}} u\theta}\times\nonumber \\
&\times \prod_{t\leq T}{\rm d}{\rho}(t){\rm d}\gamma(t){\rm d} {\hat \rho}(t)\; \delta [{\gamma}(t)- f_{\chi (t)}(\rho(t-1),y)]\nonumber\times\\&\times \prod_{t\leq T} e^{{\rm i}\hat\rho(t)\left[
	\rho(t)-\sum_{s<t}\mathcal{\hat G}(t,s)\gamma(s)+{\rm i}\sum_{s\leq T}\mathcal {\hat{\tilde C}}(t,s)\hat\rho(s)-\mathcal{\hat{\tilde B}}(t)u\right]}e^{{\rm i}\rho(t)l(t)}\label{Z_eff}
\end{align}
Here, for convenience, we have introduced 
\begin{equation}
\tilde{\rm R}_\matr A \doteq{\rm R}_{\matr A}(-\zeta)+1
\end{equation}
We next integrate out the variable $u$ in \eqref{Z_eff}. To this, we define 
\begin{equation}
\kappa(t)\doteq\frac{{\rm i}}{\sqrt{\chi}\tilde {\rm R}_\matr A}\mathcal{\hat{\tilde B}}(t)\label{kappaf}.
\end{equation}
By using the Gaussian integration formula we get
\begin{equation}
\int {\rm d}u\; e^{-\frac{\tilde{\rm R}_\matr A}{2}u^2-u\sqrt{\chi}\tilde {\rm R}_\matr A\sum_{t}\kappa(t)\hat\rho(t)}e^{\frac{{\rm i}}{\sqrt{\chi}} u\theta}=\frac{\sqrt{2\pi}}{\sqrt{\tilde {\rm R}_{\matr A}}}e^{\frac{\chi\tilde{\rm R}_\matr A}{2}(\sum_{t}\kappa(t)\hat\rho(t))^2} e^{-\frac{\theta^2}{2\chi\tilde{\rm R}_\matr A}-{\rm i}\sum_{t}\kappa(t)\hat\rho(t)\theta}.\label{gaussi}
\end{equation} 
We also re-represent the temporal couplings of $\{\hat\rho(t)\}$ via the averages of appropriate Gaussian fields. Doing so, we finally obtain 
\begin{align} 
Z_{ef}(\{l(t)\})=\int &\mathcal N(\{\phi(t)\}\vert \matr 0,\mathcal C_\phi){\rm d}\mathcal N(\theta\vert 0,q){\rm d}y\; p(y\vert \theta)\prod_{t=1}^{T}{\rm d}{\rho}(t) \times \nonumber  \\&\times\delta \left[\rho(t)-\sum_{s<t}\hat{\mathcal G}(t,s)f_{\chi(t)}(\rho(t-1),y)-\phi(t)-\kappa(t)\theta\right]e^{{\rm i}\rho(t)l(t)}.\label{Zefnew}
\end{align}
Here, we have defined $\mathcal  C_\phi\doteq2\mathcal{\hat{\tilde C}}$ and
\begin{equation}
q\doteq\left(\frac{1}{\chi \tilde {\rm R}_\matr A}-\lambda  \right)^{-1}. \label{q}
\end{equation} 
We next bypass the need for $u$ in representing the order parameters $\zeta$ and $\mathcal {B}$. This will be possible by propagating $u^2$ and $u$ through the derivative of the integral in \eqref{gaussi} with respect to $\tilde {\rm R}_\matr A$ and $\theta$, respectively. Then, it is easy to show that
\begin{align}
\zeta&=\frac{\lambda\chi}{\lambda\chi\tilde {\rm R}_\matr A-1}\label{nu}\\
\mathcal {B}(t)&=\frac{{\rm i}}{\sqrt{\chi}\tilde {\rm R}_\matr A}\mathbb E[\theta\gamma(t)]. \label{mathcalB}
\end{align} 
In particular, from \eqref{mathcalB} we write the dynamical order parameters of the effective generating functional \eqref{Zefnew} as
\begin{align}
\kappa(t)&=\frac{1}{\chi\tilde{\rm R}_\matr A^2}\sum_{n=2}^\infty c_{\matr A,n}\sum_{k=0}^{n-2}(-\zeta)^{k}\sum_{s<t}\mathbb E[\theta\gamma(s)]\mathcal G^{n-k-2}(t,s)\label{kap}\\
\mathcal{C}_\phi(t,s)&=\sum_{n=2}^\infty c_{\matr A,n}\sum_{k=0}^{n-2}(\mathcal G^k \mathcal {C}(\mathcal G^\top)^{n-2-k})(t,s)-{\chi\tilde{\rm R}_\matr A}\kappa(t)\kappa(s)\nonumber \\
&+\frac{1}{\chi\tilde{\rm R}_\matr A^2}\sum_{n=3}^\infty c_{\matr A,n}\sum_{k=0}^{n-2}(-\zeta)^{k}\sum_{l=0}^{n-k-3}\sum_{t'>t,s< s'}(\mathcal G^\top)^l(t,t')\mathbb E[\theta\gamma(t')]\mathbb E[\theta\gamma(s')]\mathcal G^{n-k-l-3}(s',s)\label{covf}.
\end{align}
Finally, notice that the response function $\mathcal G (t,s)=\mathbb E[{-\rm i}\gamma(t)\hat\rho(s)]_{Z_{ef}}$ can be written as 
\begin{equation}
\mathcal G(t,s) =\mathbb E\left[\frac{\partial \gamma (s)}{\partial \phi(\tau)}\right]_{Z_{ef}}.\label{resp}
\end{equation}
\subsection{Vanishing of memory terms}\label{vanishmemory}
We next show the memory-freeness property $\mathcal{G} =\matr 0$ which also implies $\mathcal{\hat G}= \matr 0$. Note that this leads \eqref{Zefnew} to 
\begin{equation} 
Z_{ef}(\{l(t)\})=\int \mathcal N(\{\phi(t)\}\vert \matr 0,\mathcal C_\phi){\rm d}\mathcal N(\theta\vert 0,q){\rm d}y\; p(y\vert \theta)\prod_{t=1}^{T}{\rm d}{\rho}(t)\; \delta[\rho(t)-\phi(t)-\kappa(t)\theta]e^{{\rm i}\rho(t)l(t)}.
\end{equation}

The memory-freenesss property easily follows from the fact that
\begin{equation}
\mathbb E\left[f_{\chi(t)}'(\rho(t-1),y) \right]_{Z_{ef}} = 0,\quad \forall t
\end{equation}
where $f_\chi'(\rho,y)$ stands for the derivative $\frac{\partial f_\chi(\rho,y)}{\partial \rho}$.
Specifically, note from \eqref{resp}, that
\begin{equation}
\mathcal G(t,\tau) =
\mathbb E\left[f_{\chi(t)}'(\rho(t-1),y)\frac{\partial \rho(t-1)}{\partial \phi(\tau)}\right]_{Z_{ef}}\label{resonse}.
\end{equation}
Hence, we have (see \eqref{Zefnew})
\begin{align}
\mathcal G(t,\tau) =& \sum_{\tau < s < t} \hat{\mathcal{G}}(t -1,s) \mathbb E\left[f_{\chi (s)}'(\rho(s-1),y)\frac{\partial \rho(s-1)}{\partial \phi(\tau)}\right]_{Z_{ef}}+ \underbrace{\mathbb E\left[f_{\chi(t)}'(\rho(t-1),y) \right]_{Z_{ef}}}_{0}\delta_{t-1,\tau}.
\label{resonse_it}
\end{align}
The trivial solution $\mathcal{G} = \hat{\mathcal{G}}=\matr 0$ fulfills this equation. This solution is unique given the fact that
from the definitions of $\hat{\mathcal{G}}$ and
$\mathcal{G}$, one can show that these quantities are uniquely defined recursively 
in time as expectations over the stochastic process $\{\phi(t)\}$. 

\subsection{Derivation of the relation $q=\frac 1 N {\rm tr}(\matr K)$}
We next show that the solution 
\begin{equation}
q=\frac 1 N {\rm tr}(\matr K)={\rm R}_\matr K(0)
\end{equation}
solves both \eqref{nu} and \eqref{q}. To show this, we will repeatedly use the R-transform formulation for the inverse of matrices \cite{ralf08}:
\begin{equation}
\frac{1}{{\rm R}_{\matr D}(\omega)}={\rm R}_{\matr D^{-1}}\left(-{\rm R}_{\matr D}(\omega)[1+\omega {\rm R}_{\matr D}(\omega)] \right). 
\end{equation}

First, we invoke the solution $q={\rm R}_{\matr K}(0)$ into \eqref{q} as
\begin{align}
\frac{1}{{\rm R}_{\matr K}(0)}&={\rm R}_{\matr K^{-1}}(-{\rm R}_{\matr K}(0))\\
&=\frac{1}{\chi{\rm R}_{\frac{1}{\chi}(\lambda {\bf I}+\matr K^{-1})^{-1}}(-\zeta)}-\lambda\\
&=\frac{1}{{\rm R}_{(\lambda {\bf I}+\matr K^{-1})^{-1}}(-\frac \zeta\chi)}-\lambda\\
&={\rm R}_{\matr K^{-1}}(-\underbrace{{\rm R}_{(\lambda {\bf I}+\matr K^{-1})^{-1}}(-\frac\zeta\chi)[1-\frac\zeta\chi{\rm R}_{(\lambda {\bf I}+\matr K^{-1})^{-1}}(-\frac \zeta\chi)]}_{{\rm R}_{\matr K}(0)} )+\lambda -\lambda.
\end{align}
Then, we write everything in terms of ${\rm R}_{\matr K}(0)$ as
\begin{align}
{\rm R}_{\matr K}(0)&={\rm R}_{(\lambda {\bf I}+\matr K^{-1})^{-1}}(-\frac\zeta\chi)[1-\frac\zeta\chi{\rm R}_{(\lambda {\bf I}+\matr K^{-1})^{-1}}(-\frac \zeta\chi)]\\
&=\frac{1}{\lambda +{\rm R}_{\matr K^{-1}}(-{\rm R}_{\matr K}(0))}[1-\frac \zeta\chi\frac{1}{\lambda +{\rm R}_{\matr K^{-1}}(-{\rm R}_{\matr K}(0))}]\\
&=\frac{1}{\lambda +{1}/{{\rm R}_{\matr K}(0)}}\left[1-\frac \zeta\chi\frac{1}{\lambda +{1}/{{\rm R}_{\matr K}(0)}}\right]
\end{align}
which implies that
\begin{equation}
-\frac{\zeta}{\chi}=\lambda+q\lambda^2. \label{zeta}
\end{equation}
Indeed, from \eqref{q} this is equivalent to \eqref{nu}.
\subsection{Simplification of the order parameters $\kappa(t)$ and $\mathcal C_\phi(t,s)$}

We invoke the result $\mathcal G=\matr 0$ in \eqref{kap} and get
\begin{align}
\kappa(t)&=\frac{\mathbb E[\theta\gamma(t)]}{-\zeta\chi\tilde{\rm R}_\matr A^2}\sum_{n=2}^\infty c_{\matr A,n}(-\zeta)^{n-1}\\
&=\frac{\tilde{\rm R}_\matr A-1}{-\zeta\chi\tilde{\rm R}_\matr A^2}\mathbb E[\theta\gamma(t)]\label{bson} \\
&=\frac{\kappa}{q\lambda}\mathbb E[\theta\gamma(t)]\label{son}.
\end{align}
Here, we get \eqref{son} by using the relations (see \eqref{q} and \eqref{zeta}, respectively)
\begin{align}
\zeta&=-q\lambda(1-\chi \kappa) \\
\tilde{\rm R}_\matr A&=(1-\chi\kappa)^{-1}. 
\end{align}

Second, as regards the expression \eqref{covf}, given the fact that $\mathcal G=\matr 0$  we note that 
\begin{align}
\sum_{n=3}^\infty c_{\matr A,n}\sum_{k=0}^{n-2}(-\zeta)^{k}\sum_{l=0}^{n-k-3}(\mathcal G^{n-k-3})(t,s)&
=\frac{1}{\zeta ^2}\sum_{n=3}^{\infty}c_{\matr A,n}(-\zeta)^{n-1}\\
&=\frac{{\rm R}_\matr A(-\zeta)}{\zeta^2}+\frac{\sigma^2_{\matr A}}{\zeta}.
\end{align}
Hence, we get from \eqref{covf}
\begin{align}
\mathcal C_\phi(t,s)&=\sigma_{\matr A}^2\mathbb E[\gamma(t)\gamma(s)]-\chi\tilde{\rm R}_\matr A\kappa(t)\kappa(s)+\frac{\sigma_{\matr A}^2}{{\chi\tilde {\rm R}_\matr A^2}\zeta}\mathbb E[\theta\gamma(t)]\mathbb E[\theta\gamma(s)]+\frac{\tilde{\rm R}_\matr A-1}{\chi\tilde {\rm R}_\matr A^2\zeta^2}\mathbb E[\theta\gamma(t)]\mathbb E[\theta\gamma(s)]\\
&=\sigma_{\matr A}^2\mathbb E[\gamma(t)\gamma(s)]-\chi\tilde{\rm R}_\matr A\kappa(t)\kappa(s)
+\sigma_{\matr A}^2\frac{\zeta}{\chi}\frac{\kappa(t)\kappa(s)}{\kappa^2}+\frac{\tilde{\rm R}_\matr A-1}{\chi}\frac{\kappa(t)\kappa(s)}{\kappa^2}\\
&=\sigma_{\matr A}^2\mathbb E[\gamma(t)\gamma(s)]-\sigma_{\matr A}^2(\lambda+q\lambda^2) \frac{\kappa(t)\kappa(s)}{\kappa^2}+\frac{\tilde{\rm R}_\matr A-1}{\chi}\frac{\kappa(t)\kappa(s)}{\kappa^2}-\chi\tilde{\rm R}_\matr A\kappa(t)\kappa(s)\\
&=\sigma_{\matr A}^2\mathbb E[\gamma(t)\gamma(s)]-\sigma_{\matr A}^2(\lambda+q\lambda^2)\frac{\kappa(t)\kappa(s)}{\kappa^2}+\frac{\kappa(t)\kappa(s)}{\kappa}.
\end{align} 
   
\section{Derivation of \eqref{rc} and \eqref{newrate}}\label{drc}
It follows from the recursive expression \eqref{covr} that the variances $\tau_{\phi}(t+1)\doteq \mathcal C(t+1,s+1)$ do not depend on $\mathcal C_\phi(t,s)$ for $t\neq s$ but solely on the previous variances $\tau_\phi(t)$. We also note that 
\begin{equation}
\mathcal C_{\rho}(t,s)=\mathcal C_{\phi}(t,s)+q\kappa(t)\kappa(s).\label{c_rho}
\end{equation}
\subsection{Derivation of \eqref{rc}}
Using the equations \eqref{covr} and \eqref{c_rho} in the expression \eqref{decdelta} we write 
\begin{align}
\Delta(t,s)&=-2\mathcal C_\phi(t,s)+\tau_\phi(t)+\tau_\phi(s)+q(\kappa(t)-\kappa (s))^2 \\
&=-2\sigma_\matr A^2\underbrace{\mathbb E[f_{\chi(t)}(\rho(t-1),y)f_{\chi(s)}(\rho(s-1),y)]}_{g_{t-1,s-1}[\Delta(t-1,s-1)]}+ h[\tau_\phi(t-1),\tau_\phi(s-1)]\label{rkey}
\end{align}
for an appropriately defined function $h$\footnote{Specifically, $
h[\tau_\phi(t-1),\tau_\phi(s-1)]\doteq \tau_\phi(t)+\tau_\phi(s)+q(\kappa(t)-\kappa (s))^2 -2 \frac{\kappa(t)\kappa(s)}{\kappa^2}(\kappa-\sigma^2_{\matr A}(\lambda +q\lambda^2))
	$.}.  We next define the function $g_{t,s}(x)$ explicitly by using the representation of the Gaussian density in terms of the characteristic function as
\begin{align}
g_{t,s}(x)&\doteq \frac{1}{(2\pi^2)}\int  {\rm d}\mathcal N(\theta\vert 0,q){\rm dy}{\rm d}\phi_1{\rm d}\phi_2{\rm d}k_1{\rm d}k_2\; p(y\vert \theta)\nonumber \\& \qquad \qquad\times  f_{\chi(t+1)}(\phi_1+\mathcal \kappa(t)\theta,y)f_{\chi(s+1)}(\phi_2+\mathcal \kappa(s)\theta,y)\times \nonumber \\ & \qquad \qquad \times e^{-{\rm i}k_1\phi_1-{\rm i}k_2\phi_2}e^{-\frac{1}{2}[\tau_\phi(t)k_1^2+\tau_\phi(s)k_2^2]}
e^{-\frac{1}{2}k_1k_2[\tau_\phi(t)+\tau_\phi(s)+q(\kappa(t)-\kappa (s))^2]}e^{\frac{x}{2}k_1k_2}.
\end{align}
So that we have  (for $t\neq s$)
\begin{equation}
\frac{\partial {g_{t,s}(\Delta(t,s))}}{\partial \Delta(t,s)}=-\frac{1}{2}\mathbb E [f'_{\chi(t+1)}(\rho(t),y)f'_{\chi(s+1)}(\rho(s),y)].
\end{equation}
Hence, we get 
\begin{align}
\frac{\partial \Delta(t+1,s+1)}{\partial\Delta(t,s)}&=\sigma_\matr A^2\mathbb E [f'_{\chi(t+1)}(\rho(t),y)f'_{\chi(s+1)}(\rho(s),y)]
\end{align}
where $f_\chi'(\rho,y)$ stands for the derivative $\frac{\partial f_\chi(\rho,y)}{\partial \rho}$, specifically, 
\begin{equation}
f_\chi'(\rho,y)=\frac{1}{\chi}m_\nu'(\rho,y)-1.
\end{equation}
Thus, for sufficiently large $t$ and $s$ we can expand $\Delta(t + 1,s + 1)$ around $0$ as
\begin{equation}
\Delta(t+1,s+1)=\sigma_\matr A^2\mathbb E[f'_{\chi}(\rho,y)^2]\Delta(t,s)+O(\Delta(t,s)^2).
\end{equation}
Then, we get
\begin{align}
\lim_{t,s\to\infty}\frac{\Delta(t+1,s+1)}{\Delta(t,s)}&=\lim_{t,s\to\infty}(\sigma_\matr A^2\mathbb E[f'_{\chi}(\rho,y)^2]+O(\Delta(t,s)))\nonumber \\
&=\sigma_\matr A^2\mathbb E[f'_{\chi}(\rho,y)^2].
\end{align}
This completes the derivation of \eqref{rc}. 
\subsection{Derivation of \eqref{newrate}}\label{dnewrate}
We have
\begin{align}
\Delta(t,\infty)&=-2\mathcal C_\phi(t,\infty)+\tau(t)+\kappa+q(\kappa(t)-\kappa)^2.
\end{align}
where $\mathcal C_\phi(t,\infty)\doteq \lim_{s\to \infty}\mathcal C_\phi(t,s)$. Similar to \eqref{rkey}, we then write the recursion for $\Delta(t,\infty)$ as
\begin{align}
\Delta(t+1,\infty)=-2\sigma_\matr A^2g_{t}[\Delta(t,\infty)]+h(\tau_\phi(t)).
\end{align} 
Here we have defined
\begin{align}
g_{t}(x)&\doteq \frac{1}{(2\pi^2)}\int  {\rm d}\mathcal N(\theta\vert 0,q){\rm dy}{\rm d}\phi_1{\rm d}\phi_2{\rm d}k_1{\rm d}k_2\; p(y\vert \theta)\nonumber \\& \qquad \qquad\times  f_{\chi(t+1)}(\phi_1+\mathcal \kappa(t)\theta,y)f_{\chi}(\phi_2+\mathcal \kappa\theta,y)\times \nonumber \\ & \qquad \qquad \times e^{-{\rm i}k_1\phi_1-{\rm i}k_2\phi_2}e^{-\frac{1}{2}[\tau_\phi(t)k_1^2+\kappa k_2^2]}
e^{-\frac{1}{2}k_1k_2[\tau_\phi(t)+\kappa+q(\kappa(t)-\kappa)^2]}e^{\frac{x}{2}k_1k_2}.\\
h(\tau_\phi(t))&\doteq\tau_\phi(t+1)+\kappa+q(\kappa(t+1)-\kappa )^2 -2 \frac{\kappa(t+1)}{\kappa}(\kappa-\sigma^2_{\matr A}(\lambda +q\lambda^2)).
\end{align}
In particular, we have the derivative
\begin{align}
\frac{\partial g_t[\Delta(t,\infty)]}{\partial\Delta(t,\infty)}&=-\frac{1}{2}\mathbb E [f'_{\chi(t+1)}(\phi(t)+\kappa(t)\theta,y)f'_{\chi}(\phi+\kappa\theta,y)] \label{kkey}
\end{align}
where the random variables $\phi(t)$ and $\phi$ are jointly Gaussian with zero mean and covariance $\mathcal C_\phi(t,\infty)$, and independent of $\theta$.

We are interested in the rate of the asymptotic decay 
\begin{equation}
\Delta(t,\infty) \simeq e^{t\kappa}\qquad t\to\infty.
\end{equation}
The rate is computed as 
\begin{align}
\kappa &=\lim_{t\to\infty} \ln \frac{\partial \Delta(t+1,\infty)}{\partial\Delta(t,\infty)}.\label{kappa}
\end{align}
Then, using the result \eqref{kkey} we obtain the rate as
\begin{align}
\kappa &=\lim_{t\to\infty}\ln\sigma_\matr A^2\mathbb E [f'_{\chi(t+1)}(\phi(t)+\kappa(t)\theta,y)f'_{\chi}(\phi+\kappa\theta,y)]\\
&=\ln\sigma_\matr A^2\mathbb E [f'_{\chi}(\rho,y)^2]\\
&=\ln\mu_\rho.
\end{align}
This completes the derivation of \eqref{newrate}.
\section{The saddle point argument}\label{sparg}
For convenience, let $\matr Z\doteq\matr z\matr z^\top$. Note that 
\begin{equation}
(\matr Q-\matr Z)^n=\matr Q^n+(-\matr Z)^n+\sum_{\sum_k{(i_k+j_k)}=n}\matr Q^{i_1}(-\matr Z)^{j_1}\matr Q^{i_2}(-\matr Z)^{j_2}\cdots\matr Q^{i_{n-1}}(-\matr Z)^{j_{n-1}}.
\end{equation}
Furthermore, since $\matr Z$ is rank-one, we have for $\zeta\doteq \matr z^\top\matr z$ that
\begin{align}
\matr Q^{i_1}(-\matr Z)^{j_1}\matr Q^{i_2}(-\matr Z)^{j_2}\cdots=(-1)\zeta^{j_1-1}(-1)\zeta^{j_2-1}\cdots\matr Q^{i_1}\matr Z\matr Q^{i_2}\matr Z\cdots\;.
\end{align}
Moreover, by the cyclic invariance property of the trace we have 
\begin{align}
{\rm tr}(\matr Q^{i_1}\matr Z\matr Q^{i_2}\matr Z\cdots)&=(\matr z^\top\matr Q^{i_1}\matr z)(\matr z^\top\matr Q^{i_2}\matr z)\cdots\\
&=([\mathcal{\tilde B}~~\mathcal {B}]\mathcal Q^{i_1-1} [\mathcal {B}~~\mathcal {\tilde B}]^\top)([\mathcal{\tilde B}~~\mathcal {B}]\mathcal Q^{i_2-1} [\mathcal { B}~~\mathcal {\tilde B}]^\top)\cdots 
\end{align}
For example, 
\begin{align}
[\mathcal{\tilde B}~~\mathcal B]\mathcal Q [\mathcal {B}~~\mathcal {\tilde B}]^\top&=2\mathcal{\tilde{B}} \mathcal G\mathcal B^\top+\mathcal{\tilde{B}} \mathcal C\mathcal {\tilde B}^\top+\mathcal{{B}} \mathcal {\tilde C}\mathcal {B}^\top\\&=2\mathcal{\tilde{B}} \mathcal G\mathcal B^\top+\mathcal{{B}} \mathcal {\tilde C}\mathcal {B}^\top+ {\rm SP}(\mathcal G, \mathcal C,\mathcal {\tilde C},\mathcal B,\mathcal {\tilde B},\zeta)\\
[\mathcal{\tilde B}~~\mathcal B]\mathcal Q^{n} [\mathcal {B}~~\mathcal {\tilde B}]^\top&=2\mathcal{\tilde{B}} \mathcal G^n\mathcal B^\top+\sum_{k=0}^{n-1}\mathcal{B}\mathcal G^k\mathcal {\tilde C}(\mathcal G^{\top})^{n-k-1}\mathcal {B}^\top+{\rm SP}(\mathcal G, \mathcal C,\mathcal {\tilde C},\mathcal B,\mathcal {\tilde B},\zeta)\\
([\mathcal{\tilde B}~~\mathcal B]\mathcal Q [\mathcal {B}~~\mathcal {\tilde B}]^\top)^2&={\rm SP}(\mathcal G, \mathcal C,\mathcal {\tilde C},\mathcal B,\mathcal {\tilde B},\zeta)
\end{align}
Thereby, we get
\begin{align}
{\rm tr}((\matr Q-\matr Z)^n)&={\rm tr}(\matr Q^n)+(-\zeta)^n-\sum_{i_1+i_2+j=n}(-\zeta)^{j-1}{\rm tr}(\matr Q^{i_1}\matr Z\matr Q^{i_2})+{\rm SP}(\mathcal G, \mathcal C,\mathcal {\tilde C},\mathcal B,\mathcal {\tilde B},\zeta).
\end{align} 
Again by cyclic invariance we have 
\begin{align}
&\sum_{i_1+i_2+j=n}(-\zeta)^{j-1}{\rm tr}(\matr Q^{i_1}\matr Z\matr Q^{i_2})=n\sum_{k=0}^{n-2}(-\zeta)^{k}{\rm tr}(\matr Q^{n-k-1}\matr Z)\\
&=n\sum_{k=0}^{n-2}(-\zeta)^{k}([\mathcal{\tilde B}~~\mathcal B]\mathcal Q^{n-k-2} [\mathcal {B}~~\mathcal {\tilde B}]^\top)\\
&=2n\sum_{k=0}^{n-2}(-\zeta)^{k}(\mathcal{\tilde B}\mathcal G^{n-k-2}\mathcal {B}^\top)+n(1-\delta_{n2})\sum_{k=0}^{n-2}(-\zeta)^{k}\sum_{l=0}^{n-k-3}\mathcal{B}\mathcal G^l\mathcal {\tilde C}(\mathcal G^\top)^{n-k-l-3}\mathcal {B}^\top+\nonumber \\& +{\rm SP}(\mathcal G, \mathcal C,\mathcal {\tilde C},\mathcal B,\mathcal {\tilde B},\zeta).
\end{align}
Moreover, one can show that 
\begin{equation}
{\rm tr}(\matr Q^n)=2{\rm tr}(\mathcal G^n)+ n\sum_{k=0}^{n-2}{\rm tr}(\mathcal G^k \mathcal C (\mathcal G^\top)^{n-2-k} \mathcal {\tilde C})+{\rm SP}(\mathcal G, \mathcal C,\mathcal {\tilde C},\mathcal B,\mathcal {\tilde B},\zeta).
\end{equation}
Putting everything together leads to \eqref{verygood}. 

\bibliographystyle{iopart-num}
\bibliography{mybib}
\end{document}